%% file: root.tex
\documentclass[letterpaper, 10 pt, conference]{ieeeconf}  

\IEEEoverridecommandlockouts                              

\usepackage[utf8]{inputenc} 
\usepackage[T1]{fontenc}    
\usepackage{hyperref}       
\usepackage{url}            
\usepackage{amsfonts}       
\usepackage{nicefrac}       
\usepackage{microtype}      
\usepackage{xcolor}         
\usepackage{cite}
\usepackage{multirow}

\usepackage{amsmath} 
\usepackage{amssymb}  
\usepackage{diagbox}
\usepackage{graphicx}
\usepackage{algorithm2e}
\usepackage{balance}
\usepackage{caption}
\usepackage{subcaption}
\usepackage{multirow}
\usepackage{booktabs}
\usepackage[free-standing-units=true]{siunitx}

\begin{document}

\title{
EVOPS Benchmark: Evaluation of Plane Segmentation \\ from RGBD and LiDAR Data
}

\author{Anastasiia Kornilova$^{1}$, Dmitrii Iarosh$^{2}$, Denis Kukushkin$^{1}$, Nikolai Goncharov$^{1}$, \\ Pavel Mokeev$^{2}$, Arthur Saliou$^{2}$, Gonzalo Ferrer$^{1}$
}


\twocolumn[{%
\renewcommand\twocolumn[1][]{#1}
\maketitle

\begin{center}
    \centering
    \captionsetup{type=figure}
    \includegraphics[width=0.98\textwidth]{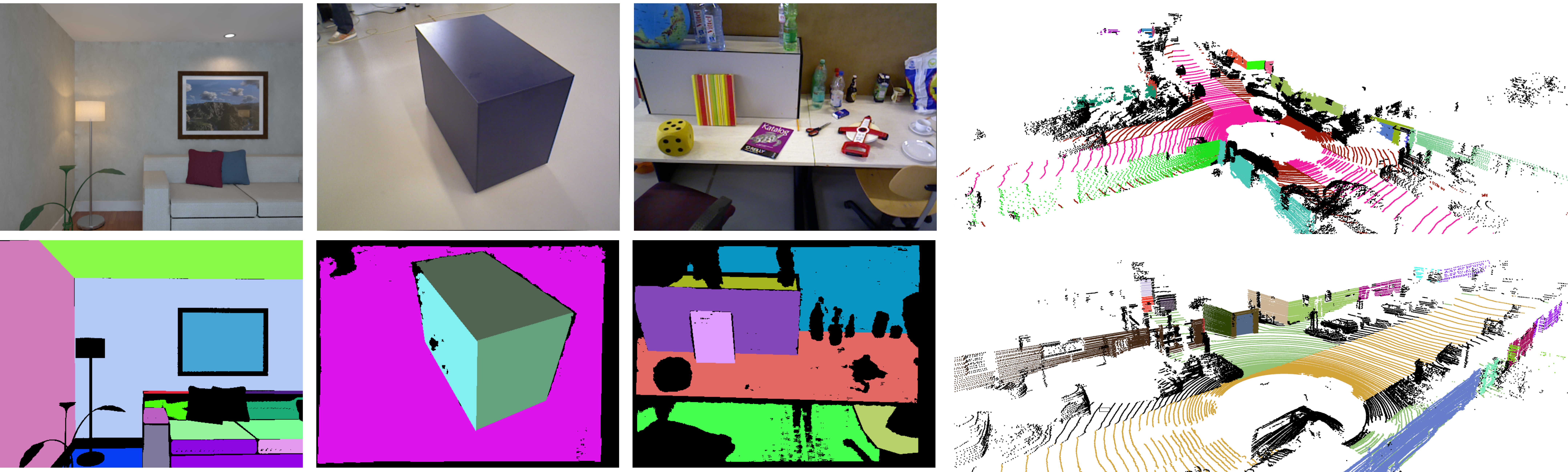}
    \captionof{figure}{Examples of annotated planes from the EVOPS dataset. The first column presents original image (top) and annotated planes mask (bottom) for ICL NUIM dataset scene, second and third~--- TUM RGDB scenes. The last column presents examples of annotated point clouds from KITTI 00 map.}
    \label{fig:teaser}
\end{center}
}]


\begin{abstract}

This paper provides the EVOPS dataset for plane segmentation from 3D data, both from RGBD images and LiDAR point clouds. We have designed two annotation methodologies (RGBD and LiDAR) running on well-known and widely-used datasets for SLAM evaluation and we have provided a complete set of benchmarking tools including point, planes and segmentation metrics.
The data includes a total number of 10k RGBD and 7K LiDAR frames over different selected scenes which consist of high quality segmented planes.
The experiments report quality of SOTA methods for RGBD plane segmentation on our annotated data. We also have provided learnable baseline for plane segmentation in LiDAR point clouds. All labeled data and benchmark tools used have been made publicly available~\url{https://evops.netlify.app/}.
\end{abstract}

\addtocounter{footnote}{1}
\footnotetext{The authors with the Skolkovo Institute of Science and Technology (Skoltech), Center for AI Technology (CAIT).
         {\tt\small {anastasiia.kornilova, g.ferrer}@skoltech.ru}}
\addtocounter{footnote}{1}
\footnotetext{The authors are with the Software Engineering Department, Saint Petersburg State University.}
\addtocounter{footnote}{-2}

\input{src/01_intro}
\input{src/02_related}

\input{src/03_dataset}
\input{src/04_metrics}

\input{src/05_eval}

\input{src/06_conclusion}


\bibliographystyle{bibtex/IEEEtran}
\bibliography{bibtex/IEEEexample.bib}

\end{document}

%% file: src/01_intro.tex
\section{Introduction}

The problem of plane segmentation of 3D data has received the attention of the robotics and computer vision (CV) community during the last years.
Multiple algorithms have reported successful results \cite{oehler2011efficient, enjarini2012planar, li2017improved,wang2010automatic, holz2011real,marriott2018plane,feng2014fast,dong2018efficient,roychoudhury2021plane,hulik2012fast,hulik2014continuous,poppinga2008fast,yang2010plane, awrangjeb2014automatic,wu2016fast,lejemble2020persistence,xu2020plane}, some of them even including their open source implementations for the community.
There is though a common denominator in all the above methods: their evaluations are based on a small subset of manually and imperfectly annotated planes or unrealistic synthetic data.
This paper aims to provide a common benchmark for plane segmentation, with evaluation tools and high quality annotated data.

Processing of 3D data, either RGBD or LiDAR, might benefit by using planes: these are geometric constraints to be exploited, quite abundant in man-made environments.
Downstream tasks could also benefit from planes, for instance, PC registration \cite{grant2013finding,ferrer2019eigen} or 3D SLAM \cite{salas2014dense,zhang2014loam,hsiao2017keyframe,liu2021balm}.
The plane segmentation becomes entangled to the quality of alignment, making these techniques more challenging to implement.
The annotations from EVOPS could be used for plane segmentation and instance segmentation to disambiguate between planes (front-end) or for trajectory estimation algorithms (back-end).

Correctly annotating planes is not straightforward: one needs to provide the accurate and complete boundaries, especially for low-quality point clouds collected by RGBD sensors or the varying density from LiDAR PCs.
Four different datasets are considered: the synthetic ICL NUIM (RGBD) \cite{handa2014icl} and TUM RGBD \cite{sturm2012tum}, and from LiDAR~--- synthetic CARLA \cite{dosovitskiy2017carla} and KITTI \cite{geiger2021kitti}.
We have developed specific methodology for annotating planes for different data modalities. In total, the dataset consist of 10k RGBD frames and 7K LiDAR observations, where the total amount of unique planes is 175 and 2196 respectively and the average number of planes per frame is 7.9 for different RGBD sequences and 70 for LiDAR.
An example of the labeled data is presented in Fig.~\ref{fig:teaser}.

Our contributions are listed below:
\begin{enumerate}
    \item Labeled RGBD and LiDAR data from popular SLAM/odometry evaluation datasets with data associations;
    \item Pip-package with metrics to measure segmentation accuracy~\footnote{https://pypi.org/project/evops/};
    \item Publicly available repository with docker images of existing open-source plane segmentation approaches on 3D data~\footnote{https://github.com/MobileRoboticsSkoltech/3D-plane-segmentation};
    \item Evaluation of existing 3D data approaches on RGBD data;
    \item Learnable baseline for plane segmentation in LiDAR point clouds~\footnote{https://github.com/MobileRoboticsSkoltech/dsnet-plane-segmentation}.
\end{enumerate}

%% file: src/02_related.tex
\section{Related work}

3D depth datasets are the {\em de facto} tool for the robotics and computer vision communities to test new algorithms.
The TUM RGBD dataset~\cite{sturm2012tum} is a good example, used for SLAM benchmarking~\cite{mur2015TumSeqUse1,mur2017TumSeqUse2}, visual odometry \cite{fontan2020TumUse1} or geometry extraction \cite{proenca2018cape}.
The ICL NUIM dataset \cite{handa2014icl} is a synthetically generated dataset, also widely popular in evaluations \cite{whelan2015elasticfusion,engel2018IclUse2}.

LiDAR is a different modality of 3D data effective in outdoors and large scales, such as the KITTI dataset \cite{geiger2021kitti},
or the KAIST dataset \cite{jeong2019complex}.
CARLA \cite{dosovitskiy2017carla} is simulator for autonomous driving research, which can be used to generate synthetic datasets.
All these are common benchmarks that the community has used and developed useful tools for. Our aim is to start our dataset on plane segmentation from this common ground.

RGBD plane segmentation algorithms became a popular topic with the advent of the first mass-produced Kinect depth cameras.
RANSAC \cite{fischler1981random} based plane detector are based on hypothesis testing and outlier filtering \cite{li2017improved, oehler2011efficient,enjarini2012planar}.
RANSAC-like variants are not suitable for real-time applications, though.

In contrast, region-based clusterization techniques are well suited for real-time requirements.
Multiple variants: split and merge \cite{wang2010automatic},
real-time normal clusterization \cite{holz2011real},
Gaussian mixture regression \cite{marriott2018plane},
agglomerative and pixel wise \cite{feng2014fast},
interleaved optimization \cite{dong2018efficient} or
flow fill  \cite{roychoudhury2021plane}.
All these papers provide a custom-based evaluation mostly by synthetic data or manually labeled small subsets of real data. The lack of a common benchmark is the main motivation of this paper.

Image segmentation from computer vision can be adapted to plane segmentation, by the depth channel \cite{hulik2012fast} or Hough transformation \cite{hulik2014continuous}.
Other problems consider planes, such as PC registration \cite{grant2013finding,ferrer2019eigen},
RGBD SLAM \cite{salas2014dense,hsiao2017keyframe,liu2021balm} or LiDAR mapping \cite{zhang2014loam}.

LiDAR constitutes a totally different modality of data for plane extraction and detection. Mainly, the density and dispersion of point is irregular, whereas RGBD provides a dense and uniform depth measurement in the image plane.
For instance, the work of Poppinga \cite{poppinga2008fast} intersects conics for plane extraction.
Other LiDAR plane segmentation approaches include
RANSAC-based \cite{yang2010plane},
segmentation of roofs \cite{awrangjeb2014automatic},
cross-line element growth \cite{wu2016fast},
multi-scale planar structure graph \cite{lejemble2020persistence} or
optimal-vector-field to detect plane intersections \cite{xu2020plane}.

On LiDAR data, the evaluation is even more scarce than in RGBD. There are two main variants: 1) evaluating plane segmentation on limited manually annotations; 2) defining the accuracy of the downstream task (ICP, SLAM), where planes are an intermediate representation.
Our dataset covers both purposes, we provide metrics for segmentation as well as the ground truth trajectories in case of direct use of planes.

%% file: src/03_dataset.tex
\section{Dataset}
This section describes the choice of datasets to be annotated for our benchmark, the procedure of semi-automated and automated data labeling, and the statistics gathered over labeled data.

\subsection{Dataset choice}
Since the benchmark dataset targets the evaluation of SLAM pipelines that embed planar structures, we considered popular odometry and SLAM evaluation datasets with 3D data. For each sensor modality (RGBD and LiDAR) we took representatives of both types of data~--- \textit{synthetic} and \textit{real}. Among RGBD datasets, we have chosen synthetic ICL NUIM~\cite{handa2014icl} and real TUM RGBD dataset~\cite{sturm2012tum}, both are often used in RGBD SLAM. Statistics on amount of labeled data is presented in Table~\ref{table:stat}.

For LiDAR modality, we have chosen open-source CARLA simulator \cite{dosovitskiy2017carla} and KITTI dataset \cite{geiger2021kitti}. 
The first one is capable of producing  huge amounts of synthetic data. The second is the standard one for LiDAR odometry/SLAM evaluation and contains statistics on a vast slice of SOTA methods. Additionally, it was extended with semantic labels in the SemanticKITTI~\cite{behley2019semanticKitti} dataset that opens new horizons for semantic-based SLAM approaches~\cite{chen2019suma++}.

For ICL NUIM dataset, a trajectory is taken from each of the  scenes~--- \emph{living\_room} and \emph{office\_room}. TUM RGBD dataset contains plenty of different scenes, we picked the following ones: \emph{fr3\_cabinet}~--- one that contains simple planar structure, \emph{fr2\_pioneer\_slam}~--- emulates indoor robot moving in hangar, \emph{fr1\_desk} and  \emph{fr3\_long\_office\_household\_validation}~--- indoor handheld scenes with short-loop closure and long-loop closure. 

For CARLA dataset, we have chosen map \emph{Town10} that corresponds to a large urban map with different types of buildings. Car trajectory and corresponding observations from trajectory poses are obtained using built-in CARLA autopilot. For KITTI dataset, we have chosen map 00~--- one of the largest KITTI sequences on an urban environment.

\begin{table}[t!]
\caption{Statistics on segmented planes from EVOPS: number of annotated observations (depth frames, LiDAR point clouds) per scene, number of all planes in the scene and average amount of annotated planes per one observation.}
\begin{center}
\begin{tabular}{| l | c | c | c |}
\hline
Scene &  Num of obs.  &  Total planes & Planes/obs \\
\hline
\text{icl/living\_room} & 1509 & 41 & 13 \\
\text{icl/office\_room} & 1510 & 32 & 13 \\
\text{tum/fr2\_pioneer} & 2890 & 30 & 3 \\
\text{tum/fr3\_long\_office} & \multirow{2}{2em}{2603} & \multirow{2}{1em}{33} & \multirow{2}{0.5em}{8} \\
\text{\_household\_validation} &   &    & \\
\text{tum/fr1\_desk} & 595 & 32 & 7 \\
\text{tum/fr3\_cabinet} & 1119 & 7 & 5 \\
\text{carla} & 2404 & 1012 & 109 \\
\text{kitti\_00} & 4541 & 1184 & 32 \\
\hline
\end{tabular}
\end{center}
\label{table:stat}
\end{table}

\subsection{Annotation}
To annotate different modalities and datasets, we use different techniques for semi-automatically data labeling and post-processing.


\textbf{RGBD modality} For labeling RGBD modality, RGB part is annotated by labelers and then automatically remapped to the depth component using the nearest corresponding timestamps. For this task we use the open-source CVAT annotation tool~\cite{cvat}, that provides label interpolation, polygon annotation and supports tracking polygons through the whole sequence. Unfortunately, this tool has restrictions for images to be labeled in one session, therefore sequences are annotated in parts and then matched together with extra processing.

\begin{figure}[t!] 
    \centering
    \includegraphics[width=.47\textwidth]{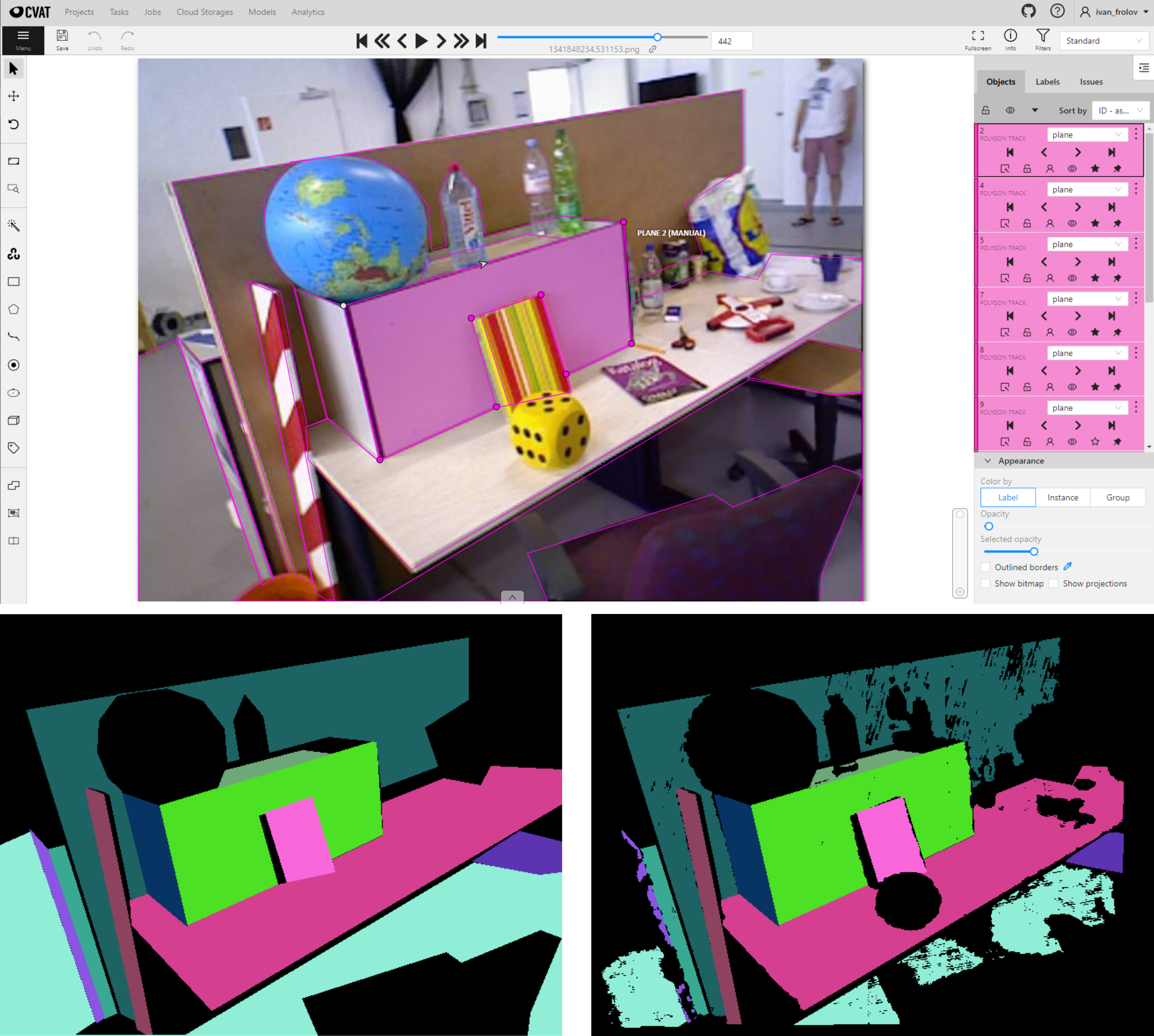}
    \caption{Stages of planes labeling on RGBD data. Top: CVAT interface with annotated planes by the labeler. Bottom-left: instance mask extracted from CVAT format. Bottom-right: post-process (final) masks using RANSAC to filter non-planar objects and infinity depth pixels.}
\label{fig:rgbd_process}
\end{figure}

Real RGBD data (TUM RGBD) contains noisy or infinite values of data at large distances and therefore could not be properly used as plane landmarks. Accordingly, on the remapping step we filter out planes that contain more than 80\% (75\% for \emph{pioneer} sequence) of infinite depth points or are located  more than 3.5 meters away from the camera, in average. To support the labeler in the time-consuming task of plane segmentation (planes that are partly overlapped by other objects) we use basic RANSAC-based plane detector from the Open3D library~\cite{open3d} in point cloud domain and then project the main segmented plane back to depth domain. An example of segmentation stages is depicted in Fig.~\ref{fig:rgbd_process}. The RANSAC parameter threshold is chosen manually for each sequence. The resulting segmentation is stored as an RGB image, where black pixels defines unsegmented points and other unique colors present segmented planes.

\textbf{LiDAR} For segmentation of LiDAR data from the KITTI dataset, we use poses from the odometry benchmark and semantic labels  from the SemanticKITTI dataset \cite{behley2019semanticKitti}. In order to effectively label the large amount of points clouds, we aggregate them into a map. As for the registration poses we used SuMa \cite{chen2019suma++} poses with loop closure that provides a more coherent map in comparison to the original KITTI ground truth poses. The resulting map is generated using every 10th point cloud with further downsampling with \SI{0.2}{\metre} voxel size and cut into submaps of 150x150\SI{}{\metre}. Every submap is manually annotated using the open-source Semantic Segmentation Editor~\cite{sse} except for road points that are taken from SemanticKITTI labels. Then, every submap is remapped back to their corresponding original observations. The final label of each point in the observation is selected as the most often label in \SI{0.2}{\metre} point vicinity. After the remap, 200 point clouds are randomly sampled among the map and verified by an operator to check the labeling correctness.

\begin{figure}[h!] 
    \centering
    \includegraphics[width=.47\textwidth]{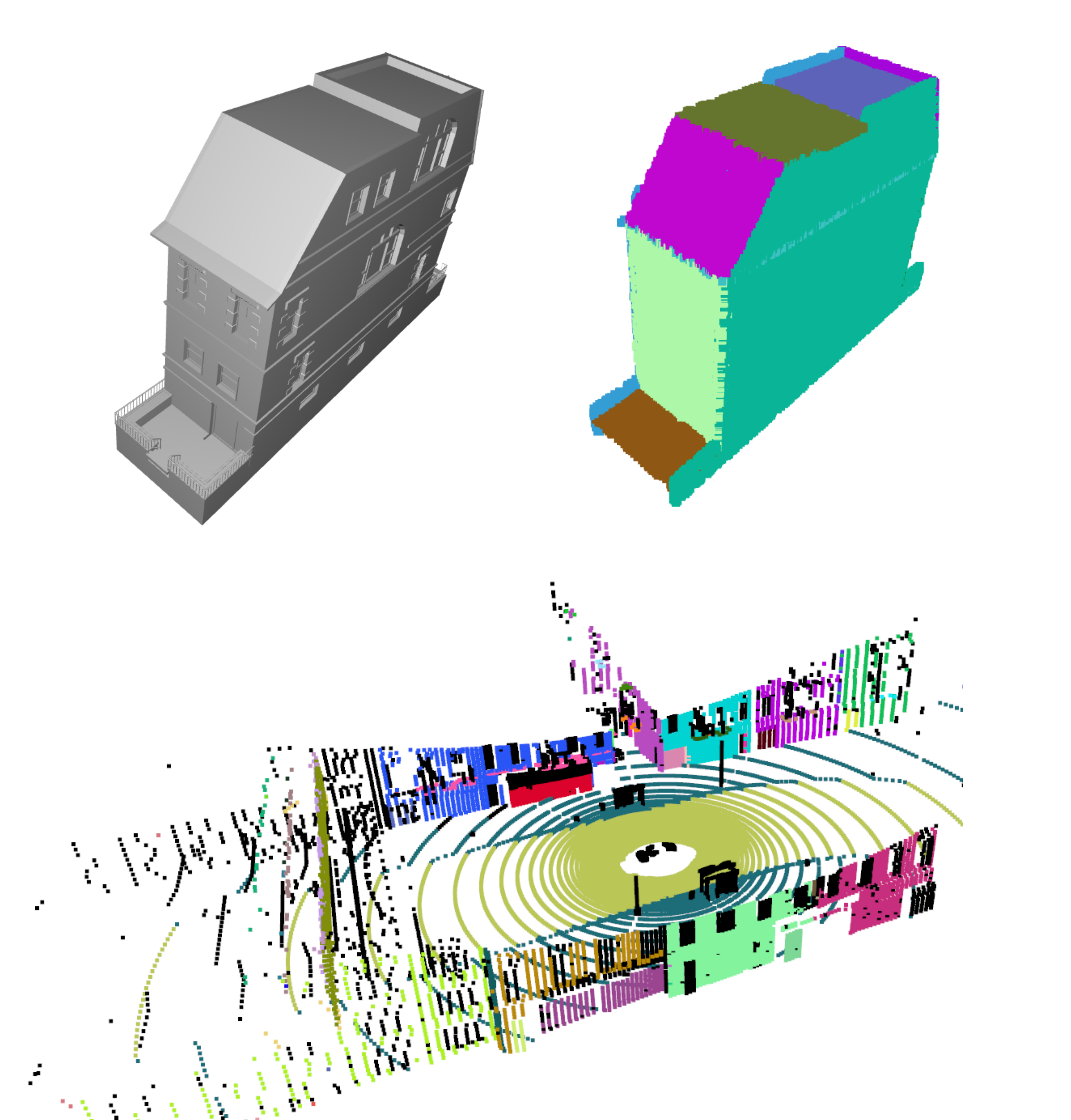}
    \caption{Stages of automatic segmentation process of CARLA simulator data. Top-left: an example of mesh presented in the simulator map. Top-right: result of our algorithm for automatic segmentation of large plane samples from this mesh. Bottom: labeled LiDAR observation segmented using built-in road and sidewalks labels from CARLA and labels from our automatic plane segmentation for building meshes.}
\label{fig:carla_process}
\end{figure}

The synthetic CARLA generator provides an API for flexible generation of LiDAR data observations and supports semantic annotation of collected point clouds. In our work, we propose a mechanism for automatic data aggregation and labeling. The main source of planar objects in outdoor scene are roads and buildings. To segment roads, we use built-in CARLA semantic labeling. To segment buildings, we implemented an automated labeling of meshes extracted from the CARLA map. To do this, we cluster mesh triangles into planar surfaces using triangle plane coefficients. Then, the sampled points from those clusters are clustered into separate subplanes to extract plane instances. Example of mesh segmentation process is presented in Fig.~\ref{fig:carla_process}. 

%% file: src/04_metrics.tex

%% file: src/05_eval.tex
\begin{figure*}[h] 
\centering
\includegraphics[width=0.95\textwidth]{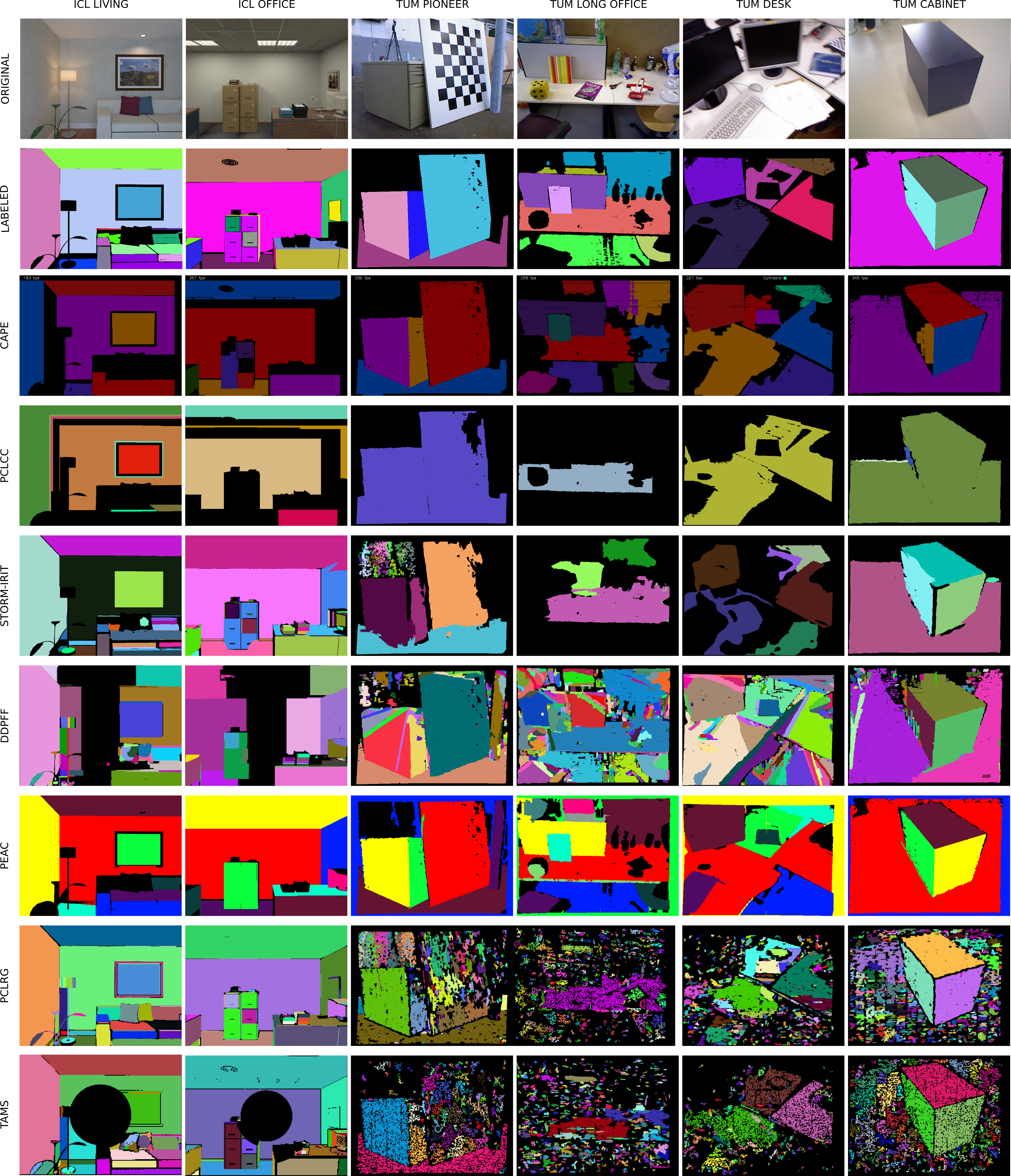}
\caption{Qualitative results of evaluation of SOTA RGBD plane segmentation methods on our dataset. Columns represent one image picked from different dataset scenes, rows~--- segmentation mask produced by considered algorithms.}
\label{fig:rgbd-qual}
\end{figure*}

\section{Evaluation}
This section provides evaluation of existing plane segmentation approaches on the EVOPS dataset. RGBD-based plane segmentation methods are much better developed and available in open-source in comparison to LiDAR-based approaches. Therefore, the first part of the evaluation considers a comparison of existing open-source methods for plane segmentation on RGBD data. The absence of open-source plane segmentation methods on LiDAR motivated us to provide a learnable baseline for this task that is described and evaluated in Sec. \ref{sec_exp_lidar}.

\subsection{Metrics}

The plane segmentation problem that our benchmark aims to solve is an instance segmentation task. For evaluation, we considered two types of metrics: \textit{plane-based}, that estimates amount of planes correctly segmented, and \textit{point-based}, that estimates amount of points from planes correctly segmented. The main difference between plane and point based metrics is that in the first one we operate only with planes, using points only to define well-overlapped planes, while in the second one we calculate all metrics using points and consider planes just like instances which are used to calculate the mean value.

For plane-based metrics
\cite{hoover1996experimental} we use: {\bf precision}, {\bf recall}, {\bf under-segmentation rate} (USR), or multiple detections of a single surface,
{\bf over-segmentation rate} (OSR), insufficient separation of multiple surfaces, 
{\bf missed}, segmentation fails to find a plane (false negative) and
{\bf noise} supposes a non-existing plane (false positive). For point-based metrics we considered popular instance segmentation metrics {\bf intersection over union} (IoU), {\bf DICE} \cite{dice1945measures}.

As all metrics are based on the intersection of the predicted and ground truth plane instances, the problem of pairing planes from predictions and ground truth is crucial. In point-based metrics it is solved by comparing all possible variations and choosing one with the maximum value of the calculated metric. However, plane-based metrics use a threshold to define that planes are overlapped enough to be taken as a pair of prediction and ground truth. For synthetic dataset (ICL NUIM) this threshold is 80\%, for real dataset (TUM RGBD and KITTI) it is 50\%.

All implemented metrics are available in the form of pip-package.


\subsection{RGBD}
For RGBD-based benchmark, we consider all publicly available implementations to the best of our knowledge. Our evaluation list includes the following algorithms: CAPE \cite{proenca2018cape}, PEAC~\cite{PEAC}, DDFPP~\cite{DDPFF}, Storm-irit\cite{STORM}, TAMS, Region Growing (PCLRG)~\cite{pclrg} from PCL library~\cite{rusu2011pcl}, Connected Components from PCL library (PCLCC). Their docker versions are available in our repository~\footnote{https://github.com/MobileRoboticsSkoltech/3D-plane-segmentation}.

Using metrics from our package, we evaluate the listed methods on all RGBD scenes from our datasets. Qualitative results for each algorithm on all dataset scenes are depicted in Fig.~\ref{fig:rgbd-qual}. Statistics on all metrics is presented in Tab.~\ref{table:eval_rgbd}.

Obtained qualitative and quantitative results highlight two leaders: CAPE and PEAC algorithms. Whereas on synthetic data other methods are close to them, especially from precision, IoU and DICE point of view, the gap on real TUM data is more tangible. On synthetic data, TAMS, PCLRG and DDPFF are close to favorites, whereas on real data, they are inclined to oversegmentation and noise (false negative) that could be crucial for SLAM pipelines. PCLCC, in contrary, provides more undersegmented results detecting only one huge plane in the whole scene that also limits its usage as detector for landmark in SLAM. Considering CAPE and PEAC in more details, it could be noticed that PEAC provides higher recall with almost the same precision that means that it detects more correct planes and therefore more efficient.


Additionally, we evaluate performance of every method. Results are presented in Tab.~\ref{table:performance}. For evaluation, machine with the following characteristics is used: 12-Core AMD Ryzen 9 3900X, 32 GB memory, NVIDIA GeForce RTX 2080 Ti. Among all methods, CAPE, DDPFF, PEAC provides performance with more than 1 Hz that makes them suitable for real-time SLAM. PEAC has the best performance, providing almost 5 Hz frequency.






\begin{table}[]
\caption{Evaluation results of plane segmentation methods for RGBD data}
\label{table:eval_rgbd}
\begin{tabular}{ccccccccc}
\midrule
Metric & \rotatebox[origin=c]{90}{Precision} & \rotatebox[origin=c]{90}{Recall} & \rotatebox[origin=c]{90}{USR}  & \rotatebox[origin=c]{90}{OSR}  & \rotatebox[origin=c]{90}{Missed} & \rotatebox[origin=c]{90}{Noise} & \rotatebox[origin=c]{90}{IoU}  & \rotatebox[origin=c]{90}{DICE} \\ \\
\midrule
\multicolumn{9}{c}{ICL}                                           \\
\midrule 
CAPE   & \textbf{0.55}  & 0.23 & 0.80 & 0.29 & 0.63  & \textbf{0.31}  & \textbf{0.64} & \textbf{0.70} \\
DDPFF  & 0.10  & 0.34 & 0.27 & 0.83 & 0.65  & 0.89  & 0.16 & 0.20 \\
PCLCC  & 0.31  & 0.17 & 0.47 & \textbf{0.12} & 0.82  & 0.63  & 0.52 & 0.60 \\
PCLRG  & 0.21  & \textbf{0.69} & \textbf{0.17} & 0.64 & \textbf{0.30}  & 0.78  & 0.25 & 0.28 \\
PEAC   & 0.50  & 0.38 & 0.74 & 0.57 & 0.61  & 0.49  & \textbf{0.64} & \textbf{0.70} \\
TAMS   & 0.24  & 0.57 & 0.28 & 0.74 & 0.42  & 0.75  & 0.28 & 0.31 \\
\midrule
\multicolumn{9}{c}{TUM}                                           \\
\midrule 
CAPE   & \textbf{0.27}  & 0.24 & 0.69 & 0.50 & 0.75  & 0.72  & \textbf{0.53} & \textbf{0.62} \\
DDPFF  & 0.01  & 0.10 & \textbf{0.07} & 0.99 & 0.89  & 0.99  & 0.01 & 0.02 \\
PCLCC  & 0.06  & 0.03 & 0.43 & \textbf{0.18} & 0.96  & \textbf{0.63}  & 0.23 & 0.30 \\
PCLRG  & 0.01  & 0.28 & 0.03 & 0.99 & 0.71  & 0.99  & 0.01 & 0.01 \\
PEAC   & 0.22  & \textbf{0.52} & 0.55 & 0.95 & \textbf{0.47}  & 0.77  & 0.30 & 0.36 \\
TAMS   & 0.01  & 0.01 & 0.04 & 0.98 & 0.99  & 0.99  & 0.01 & 0.01 \\
\midrule
\end{tabular}
\end{table}

\begin{table}[htpb]
\caption{Algorithms performance}
\begin{tabular}{cccccc}
\midrule
 \rotatebox[origin=c]{90}{CAPE} & \rotatebox[origin=c]{90}{DDPFF} & \rotatebox[origin=c]{90}{PCLCC}  & \rotatebox[origin=c]{90}{PCLRG}  & \rotatebox[origin=c]{90}{PEAC} & \rotatebox[origin=c]{90}{TAMS} \\
\midrule
\multicolumn{6}{c}{Mean time per frame (msec)}    
\\
\midrule
    472.74  & 418.22 & 20607.40 & 20651.36 & \textbf{207.08}  & 20607.40 \\
 \midrule
\multicolumn{6}{c}{Std time per frame  (msec)} \\
 \midrule

    2.01  & 1.58 & 2.58 & 114.92 & 1.83  & 68.01  \\
\midrule
\end{tabular}
\label{table:performance}
\end{table}

\subsection{LiDAR}\label{sec_exp_lidar}
Sparsity and irregularity of LiDAR data is one of the main reasons for the lack of classical methods in  plane segmentation. Learnable approaches could fill this gap by having better semantic understanding of the scene. To the best of our knowledge, there are no learnable approaches for task of plane segmentation from LiDARs. In this part we provide a baseline based on our dataset for the future developing of this direction. 

\textbf{Baseline choice.} Among the available approaches for instance segmentation task on 3D data, formulation of our problem as \textit{panoptic segmentation task} fits plane segmentation problem better in comparison to popular 3D object detection task, because in that case bounding boxes should deal with outliers in them. To choose a baseline, we follow panoptic segmentation benchmarks SemanticKITTI~\cite{behley2019semanticKitti} and nuScenes~\cite{caesar2020nuscenes}. Among methods proposed in them, DSNet \cite{hong2021lidar} has one of the best performances for this task and open-source. It uses a combination of cylindrical voxel partition and sparse convolution, which is the best we can do to compensate uneven points distribution without 2D domain mappings. Instance segmentation regresses object centers and clusters them, instead of regressing bounding boxes. It has two branches, semantic branch and instance branch, all of them use a common encoding.

\textbf{Configuration.} For training the semantic part, all planar objects of this dataset are assigned with label 1 (planar objects), the rest with label 0 (non-planar objects). The training resembles a basic semantic segmentation routine: weights for DSNet network, pre-trained on SemanticKITTI, are taken and fine-tuned on EVOPS LiDAR dataset by optimizing the cross-entropy loss. Optimal batch size and start learning rate was found to be 2 and 1e-3 respectively. The training is scheduled with exponential learning rate with multiplicative factor of 0.98. The dataset is split into train (first 75\% scenes) and test (last 20\% scenes), with dropping 5 \% of intermediate scenes to ensure that the metrics are safe from data leakage. On map 00 from KITTI dataset, that is taken into EVOPS benchmark, such split guarantees that train and test set are not intersected. The instance part is trained by freezing the semantic network weights along with the encoder, and training the instance branch. The instance branch is composed of center regression and dynamic shift clustering. The setup for the training was the same in the DSNet paper.


\textbf{Results.} Evaluation of baseline on validation set is presented in Table~\ref{table:lidar_res}.
In comparison with RGBD, the overall statistics are inferior, however segmentation on PC is a more challenging problem. The benchmarking code is made publicly available in our repository.

\begin{table}[]
\caption{Evaluation results of learnable baseline for plane segmentation on LiDAR data}
\label{table:lidar_res}
\begin{tabular}{ccccccccc}
\midrule
Metric & \rotatebox[origin=c]{90}{Precision} & \rotatebox[origin=c]{90}{Recall} & \rotatebox[origin=c]{90}{USR}  & \rotatebox[origin=c]{90}{OSR}  & \rotatebox[origin=c]{90}{Missed} & \rotatebox[origin=c]{90}{Noise} & \rotatebox[origin=c]{90}{IoU}  & \rotatebox[origin=c]{90}{DICE} \\ \\
\midrule 
DSNet & 0.04 & 0.13 & 0.22 & 0.69 & 0.86 & 0.95 & 0.31 & 0.22 \\
\midrule
\end{tabular}
\end{table}

%% file: src/06_conclusion.tex
\section{Conclusion}

In this work, we have proposed the EVOPS dataset, which is composed of different sequences from some of the most popular RGBD (ICL, TUM) and LiDAR (KITTI, CARLA) datasets.
In total, the amount of annotated frames is 10k for RGBD and 7k for LiDAR, where the amount of unique planes is 175 and 2196 correspondingly.

We have designed a methodology for semi-automatically label a high volume of data and as a result, these annotations could be used as a benchmark for plane segmentation, where we provide ready-to-use tools for evaluation and processing, including common metrics to be used by the community in a pip package.


%% file: root.bbl
\begin{thebibliography}{10}
\providecommand{\url}[1]{#1}
\csname url@rmstyle\endcsname
\providecommand{\newblock}{\relax}
\providecommand{\bibinfo}[2]{#2}
\providecommand\BIBentrySTDinterwordspacing{\spaceskip=0pt\relax}
\providecommand\BIBentryALTinterwordstretchfactor{4}
\providecommand\BIBentryALTinterwordspacing{\spaceskip=\fontdimen2\font plus
\BIBentryALTinterwordstretchfactor\fontdimen3\font minus
  \fontdimen4\font\relax}
\providecommand\BIBforeignlanguage[2]{{%
\expandafter\ifx\csname l@#1\endcsname\relax
\typeout{** WARNING: IEEEtran.bst: No hyphenation pattern has been}%
\typeout{** loaded for the language `#1'. Using the pattern for}%
\typeout{** the default language instead.}%
\else
\language=\csname l@#1\endcsname
\fi
#2}}

\bibitem{oehler2011efficient}
B.~Oehler, J.~Stueckler, J.~Welle, D.~Schulz, and S.~Behnke, ``Efficient
  multi-resolution plane segmentation of 3d point clouds,'' in
  \emph{International Conference on Intelligent Robotics and
  Applications}.\hskip 1em plus 0.5em minus 0.4em\relax Springer, 2011, pp.
  145--156.

\bibitem{enjarini2012planar}
B.~Enjarini and A.~Gr{\"a}ser, ``Planar segmentation from depth images using
  gradient of depth feature,'' in \emph{IEEE/RSJ International Conference on
  Intelligent Robots and Systems}, 2012, pp. 4668--4674.

\bibitem{li2017improved}
L.~Li, F.~Yang, H.~Zhu, D.~Li, Y.~Li, and L.~Tang, ``An improved ransac for 3d
  point cloud plane segmentation based on normal distribution transformation
  cells,'' \emph{Remote Sensing}, vol.~9, no.~5, p. 433, 2017.

\bibitem{wang2010automatic}
M.~Wang and Y.-H. Tseng, ``Automatic segmentation of lidar data into coplanar
  point clusters using an octree-based split-and-merge algorithm,''
  \emph{Photogrammetric Engineering \& Remote Sensing}, vol.~76, no.~4, pp.
  407--420, 2010.

\bibitem{holz2011real}
D.~Holz, S.~Holzer, R.~B. Rusu, and S.~Behnke, ``Real-time plane segmentation
  using rgb-d cameras,'' in \emph{Robot Soccer World Cup}.\hskip 1em plus 0.5em
  minus 0.4em\relax Springer, 2011, pp. 306--317.

\bibitem{marriott2018plane}
R.~T. Marriott, A.~Pashevich, and R.~Horaud, ``Plane-extraction from depth-data
  using a gaussian mixture regression model,'' \emph{Pattern Recognition
  Letters}, vol. 110, pp. 44--50, 2018.

\bibitem{feng2014fast}
C.~Feng, Y.~Taguchi, and V.~R. Kamat, ``Fast plane extraction in organized
  point clouds using agglomerative hierarchical clustering,'' in \emph{IEEE
  International Conference on Robotics and Automation (ICRA)}, 2014, pp.
  6218--6225.

\bibitem{dong2018efficient}
Z.~Dong, B.~Yang, P.~Hu, and S.~Scherer, ``An efficient global energy
  optimization approach for robust 3d plane segmentation of point clouds,''
  \emph{ISPRS Journal of Photogrammetry and Remote Sensing}, vol. 137, pp.
  112--133, 2018.

\bibitem{roychoudhury2021plane}
A.~Roychoudhury, M.~Missura, and M.~Bennewitz, ``Plane segmentation in
  organized point clouds using flood fill,'' in \emph{IEEE International
  Conference on Robotics and Automation (ICRA)}, 2021, pp. 13\,532--13\,538.

\bibitem{hulik2012fast}
R.~Hulik, V.~Beran, M.~Spanel, P.~Krsek, and P.~Smrz, ``Fast and accurate plane
  segmentation in depth maps for indoor scenes,'' in \emph{IEEE/RSJ
  International Conference on Intelligent Robots and Systems}, 2012, pp.
  1665--1670.

\bibitem{hulik2014continuous}
R.~Hulik, M.~Spanel, P.~Smrz, and Z.~Materna, ``Continuous plane detection in
  point-cloud data based on 3d hough transform,'' \emph{Journal of visual
  communication and image representation}, vol.~25, no.~1, pp. 86--97, 2014.

\bibitem{poppinga2008fast}
J.~Poppinga, N.~Vaskevicius, A.~Birk, and K.~Pathak, ``Fast plane detection and
  polygonalization in noisy 3d range images,'' in \emph{IEEE/RSJ International
  Conference on Intelligent Robots and Systems}.\hskip 1em plus 0.5em minus
  0.4em\relax IEEE, 2008, pp. 3378--3383.

\bibitem{yang2010plane}
M.~Y. Yang and W.~F{\"o}rstner, ``Plane detection in point cloud data,'' in
  \emph{Proceedings of the 2nd int conf on machine control guidance, Bonn},
  vol.~1, 2010, pp. 95--104.

\bibitem{awrangjeb2014automatic}
M.~Awrangjeb and C.~S. Fraser, ``Automatic segmentation of raw lidar data for
  extraction of building roofs,'' \emph{Remote Sensing}, vol.~6, no.~5, pp.
  3716--3751, 2014.

\bibitem{wu2016fast}
T.~Wu, X.~Hu, and L.~Ye, ``Fast and accurate plane segmentation of airborne
  lidar point cloud using cross-line elements,'' \emph{Remote Sensing}, vol.~8,
  no.~5, p. 383, 2016.

\bibitem{lejemble2020persistence}
T.~Lejemble, C.~Mura, L.~Barthe, and N.~Mellado, ``Persistence analysis of
  multi-scale planar structure graph in point clouds,'' in \emph{Computer
  Graphics Forum}, vol.~39, no.~2, 2020, pp. 35--50.

\bibitem{xu2020plane}
S.~Xu, R.~Wang, H.~Wang, and R.~Yang, ``Plane segmentation based on the
  optimal-vector-field in lidar point clouds,'' \emph{IEEE Transactions on
  Pattern Analysis and Machine Intelligence}, vol.~43, no.~11, pp. 3991--4007,
  2020.

\bibitem{grant2013finding}
W.~S. Grant, R.~C. Voorhies, and L.~Itti, ``Finding planes in lidar point
  clouds for real-time registration,'' in \emph{IEEE/RSJ International
  Conference on Intelligent Robots and Systems}.\hskip 1em plus 0.5em minus
  0.4em\relax IEEE, 2013, pp. 4347--4354.

\bibitem{ferrer2019eigen}
G.~Ferrer, ``Eigen-factors: Plane estimation for multi-frame and
  time-continuous point cloud alignment,'' in \emph{2019 IEEE/RSJ International
  Conference on Intelligent Robots and Systems (IROS)}, 2019, pp. 1278--1284.

\bibitem{salas2014dense}
R.~F. Salas-Moreno, B.~Glocken, P.~H. Kelly, and A.~J. Davison, ``Dense planar
  slam,'' in \emph{IEEE international symposium on mixed and augmented reality
  (ISMAR)}, 2014, pp. 157--164.

\bibitem{zhang2014loam}
J.~Zhang and S.~Singh, ``Loam: Lidar odometry and mapping in real-time.'' in
  \emph{Robotics: Science and Systems}, vol.~2, no.~9, 2014, pp. 1--9.

\bibitem{hsiao2017keyframe}
M.~Hsiao, E.~Westman, G.~Zhang, and M.~Kaess, ``Keyframe-based dense planar
  slam,'' in \emph{IEEE International Conference on Robotics and Automation
  (ICRA)}, 2017, pp. 5110--5117.

\bibitem{liu2021balm}
Z.~Liu and F.~Zhang, ``Balm: Bundle adjustment for lidar mapping,'' \emph{IEEE
  Robotics and Automation Letters}, vol.~6, no.~2, pp. 3184--3191, 2021.

\bibitem{handa2014icl}
A.~Handa, T.~Whelan, J.~McDonald, and A.~Davison, ``A benchmark for {RGB-D}
  visual odometry, {3D} reconstruction and {SLAM},'' in \emph{IEEE Intl. Conf.
  on Robotics and Automation, ICRA}, Hong Kong, China, May 2014.

\bibitem{sturm2012tum}
J.~Sturm, N.~Engelhard, F.~Endres, W.~Burgard, and D.~Cremers, ``A benchmark
  for the evaluation of rgb-d slam systems,'' in \emph{Proc. of the
  International Conference on Intelligent Robot Systems (IROS)}, Oct. 2012.

\bibitem{dosovitskiy2017carla}
A.~Dosovitskiy, G.~Ros, F.~Codevilla, A.~Lopez, and V.~Koltun, ``{CARLA}: {An}
  open urban driving simulator,'' in \emph{Proceedings of the 1st Annual
  Conference on Robot Learning}, 2017, pp. 1--16.

\bibitem{geiger2021kitti}
A.~Geiger, P.~Lenz, and R.~Urtasun, ``Are we ready for autonomous driving? the
  kitti vision benchmark suite,'' in \emph{Conference on Computer Vision and
  Pattern Recognition (CVPR)}, 2012.

\bibitem{mur2015TumSeqUse1}
R.~Mur-Artal, J.~M.~M. Montiel, and J.~D. Tardós, ``{ORB-SLAM: A Versatile and
  Accurate Monocular SLAM System},'' \emph{IEEE Transactions on Robotics},
  vol.~31, no.~5, pp. 1147--1163, 2015.

\bibitem{mur2017TumSeqUse2}
R.~Mur-Artal and J.~D. Tardós, ``Orb-slam2: An open-source slam system for
  monocular, stereo, and rgb-d cameras,'' \emph{IEEE Transactions on Robotics},
  vol.~33, no.~5, pp. 1255--1262, 2017.

\bibitem{fontan2020TumUse1}
A.~Fontan, J.~Civera, and R.~Triebel, ``{Information-Driven Direct RGB-D
  Odometry},'' in \emph{Proceedings of the IEEE/CVF Conference on Computer
  Vision and Pattern Recognition}, 06 2020, pp. 4928--4936.

\bibitem{proenca2018cape}
P.~F. Proença and Y.~Gao, ``Fast cylinder and plane extraction from depth
  cameras for visual odometry,'' in \emph{IEEE/RSJ International Conference on
  Intelligent Robots and Systems (IROS)}, 2018, pp. 6813--6820.

\bibitem{whelan2015elasticfusion}
T.~Whelan, S.~Leutenegger, R.~Salas-Moreno, B.~Glocker, and A.~Davison,
  ``Elasticfusion: Dense slam without a pose graph.''\hskip 1em plus 0.5em
  minus 0.4em\relax Robotics: Science and Systems, 2015.

\bibitem{engel2018IclUse2}
J.~Engel, V.~Koltun, and D.~Cremers, ``Direct sparse odometry,'' \emph{IEEE
  Transactions on Pattern Analysis and Machine Intelligence}, vol.~40, no.~3,
  pp. 611--625, 2018.

\bibitem{jeong2019complex}
J.~Jeong, Y.~Cho, Y.-S. Shin, H.~Roh, and A.~Kim, ``Complex urban dataset with
  multi-level sensors from highly diverse urban environments,'' \emph{The
  International Journal of Robotics Research}, vol.~38, no.~6, pp. 642--657,
  2019.

\bibitem{fischler1981random}
M.~A. Fischler and R.~C. Bolles, ``Random sample consensus: a paradigm for
  model fitting with applications to image analysis and automated
  cartography,'' \emph{Communications of the ACM}, vol.~24, no.~6, pp.
  381--395, 1981.

\bibitem{behley2019semanticKitti}
J.~Behley, M.~Garbade, A.~Milioto, J.~Quenzel, S.~Behnke, C.~Stachniss, and
  J.~Gall, ``{SemanticKITTI: A Dataset for Semantic Scene Understanding of
  LiDAR Sequences},'' in \emph{Proc. of the IEEE/CVF International Conf.~on
  Computer Vision (ICCV)}, 2019.

\bibitem{chen2019suma++}
X.~Chen, A.~Milioto, E.~Palazzolo, P.~Giguere, J.~Behley, and C.~Stachniss,
  ``Suma++: Efficient lidar-based semantic slam,'' in \emph{IEEE/RSJ
  International Conference on Intelligent Robots and Systems (IROS)}, 2019, pp.
  4530--4537.

\bibitem{cvat}
\BIBentryALTinterwordspacing
B.~Sekachev \emph{et~al.}, ``opencv/cvat: v1.1.0,'' Aug. 2020. [Online].
  Available: \url{https://doi.org/10.5281/zenodo.4009388}
\BIBentrySTDinterwordspacing

\bibitem{open3d}
Q.-Y. Zhou, J.~Park, and V.~Koltun, ``{Open3D}: {A} modern library for {3D}
  data processing,'' \emph{arXiv:1801.09847}, 2018.

\bibitem{sse}
\BIBentryALTinterwordspacing
{Hitachi Automotive and Industry Laboratory}.
  \BIBforeignlanguage{english}{Semantic segmentation editor}. [Online].
  Available:
  \url{https://github.com/Hitachi-Automotive-And-Industry-Lab/semantic-segmentation-editor}
\BIBentrySTDinterwordspacing

\bibitem{hoover1996experimental}
A.~Hoover, G.~Jean-Baptiste, X.~Jiang, P.~J. Flynn, H.~Bunke, D.~B. Goldgof,
  K.~Bowyer, D.~W. Eggert, A.~Fitzgibbon, and R.~B. Fisher, ``An experimental
  comparison of range image segmentation algorithms,'' \emph{IEEE transactions
  on pattern analysis and machine intelligence}, vol.~18, no.~7, pp. 673--689,
  1996.

\bibitem{dice1945measures}
L.~R. Dice, ``Measures of the amount of ecologic association between species,''
  \emph{Ecology}, vol.~26, no.~3, pp. 297--302, 1945.

\bibitem{PEAC}
C.~Feng, Y.~Taguchi, and V.~R. Kamat, ``Fast plane extraction in organized
  point clouds using agglomerative hierarchical clustering,'' \emph{2014 IEEE
  International Conference on Robotics and Automation (ICRA)}, pp. 6218--6225,
  2014.

\bibitem{DDPFF}
A.~Roychoudhury, M.~Missura, and M.~Bennewitz, ``Plane segmentation using
  depth-dependent flood fill,'' in \emph{2021 IEEE/RSJ International Conference
  on Intelligent Robots and Systems (IROS)}.\hskip 1em plus 0.5em minus
  0.4em\relax IEEE, pp. 2210--2216.

\bibitem{STORM}
T.~Lejemble, C.~Mura, L.~Barthe, and N.~Mellado, ``Persistence analysis of
  multi-scale planar structure graph in point clouds,'' in \emph{Computer
  Graphics Forum}, vol.~39, no.~2.\hskip 1em plus 0.5em minus 0.4em\relax Wiley
  Online Library, 2020, pp. 35--50.

\bibitem{pclrg}
\BIBentryALTinterwordspacing
PCL. \BIBforeignlanguage{english}{Region growing segmentation (2016)}.
  [Online]. Available:
  \url{https://pcl.readthedocs.io/projects/tutorials/en/master/region_growing_segmentation.html#region-growing-segmentation}
\BIBentrySTDinterwordspacing

\bibitem{rusu2011pcl}
R.~B. Rusu and S.~Cousins, ``3d is here: Point cloud library (pcl),'' in
  \emph{2011 IEEE international conference on robotics and automation}, 2011,
  pp. 1--4.

\bibitem{caesar2020nuscenes}
H.~Caesar, V.~Bankiti, A.~H. Lang, S.~Vora, V.~E. Liong, Q.~Xu, A.~Krishnan,
  Y.~Pan, G.~Baldan, and O.~Beijbom, ``nuscenes: A multimodal dataset for
  autonomous driving,'' in \emph{Proceedings of the IEEE/CVF conference on
  computer vision and pattern recognition}, 2020, pp. 11\,621--11\,631.

\bibitem{hong2021lidar}
F.~Hong, H.~Zhou, X.~Zhu, H.~Li, and Z.~Liu, ``Lidar-based panoptic
  segmentation via dynamic shifting network,'' in \emph{Proceedings of the
  IEEE/CVF Conference on Computer Vision and Pattern Recognition}, 2021, pp.
  13\,090--13\,099.

\end{thebibliography}
